\documentclass[letterpaper, 10 pt, conference]{ieeeconf}  
\IEEEoverridecommandlockouts                              
\usepackage{graphicx}
\usepackage{amsmath}
\usepackage{amssymb}
\usepackage{booktabs}
\usepackage{algpseudocode}
\usepackage{algorithm}
\newcommand{\argmax}{\operatornamewithlimits{argmax}}

\usepackage[pagebackref,breaklinks,colorlinks]{hyperref}

\graphicspath{{figures/}}

\begin{document}

\title{3P-LLM: Probabilistic Path Planning using Large Language Model for Autonomous Robot Navigation}

\author{Ehsan Latif\\
School of Computing\\
The University of Georgia\\
{\tt\small ehsan.latif@uga.edu}
}
\maketitle

\begin{abstract}
  Much worldly semantic knowledge can be encoded in large language models (LLMs). Such information could be of great use to robots that want to carry out high-level, temporally extended commands stated in natural language. However, the lack of real-world experience that language models have is a key limitation that makes it challenging to use them for decision-making inside a particular embodiment. This research assesses the feasibility of using LLM (GPT-3.5-turbo chatbot by OpenAI) for robotic path planning. The shortcomings of conventional approaches to managing complex environments and developing trustworthy plans for shifting environmental conditions serve as the driving force behind the research. Due to the sophisticated natural language processing abilities of LLM, the capacity to provide effective and adaptive path-planning algorithms in real-time, great accuracy, and few-shot learning capabilities, GPT-3.5-turbo is well suited for path planning in robotics. In numerous simulated scenarios, the research compares the performance of GPT-3.5-turbo with that of state-of-the-art path planners like Rapidly Exploring Random Tree (RRT) and A*. We observed that GPT-3.5-turbo is able to provide real-time path planning feedback to the robot and outperforms its counterparts. This paper establishes the foundation for LLM-powered path planning for robotic systems.
\end{abstract}

\section{Introduction}
\label{sec:intro}

\section{Introduction}
\label{sec:intro}
The landscape of robotics is continually evolving, ushering in an era of advanced robots capable of performing tasks in complex and dynamic settings\cite{latif2024communication}. One of the paramount challenges in this field is the efficient and effective path planning for robots, necessitating a synergy between motion planning and perception algorithms to devise safe and optimal motion strategies. The literature has seen the emergence of potent methodologies in the realm of path planning, including graph optimization path planning \cite{dang2019graph,latif2022dgorl, latif2023gprl}, heuristic-based strategies \cite{ferguson2005guide}, and the exploration of rapidly exploring random trees (RRTs) \cite{latif2022multi, latif2023seal,latif2023communication}. Despite these advances, conventional path-planning frameworks often falter in navigating complex environments and in formulating dependable strategies amidst fluctuating environmental conditions \cite{karur2021survey,latif2023instantaneous, latif2023intersection}.

The advent of large language models (LLMs) has heralded new capabilities, from responding to queries and generating complex textual responses to engaging in diverse conversational topics. This raises the intriguing possibility of robots leveraging the vast troves of knowledge encapsulated in such models to execute complex real-world tasks. The crucial question is: \textit{how can physical entities harness and apply the knowledge from LLMs to tasks in the tangible world?}

Employing GPT-3.5-turbo for robotic path planning presents a promising avenue to circumvent the limitations inherent in traditional methodologies. GPT-3.5-turbo's advanced natural language processing prowess positions it as an ideal candidate for tackling the multifaceted and dynamic challenges in robotics. Moreover, its capability to furnish efficient and adaptable path-planning solutions in real-time lays a solid groundwork for future explorations in this domain.

\begin{figure}[t]
\centering
\begin{center}
 \includegraphics[width=0.97\columnwidth]{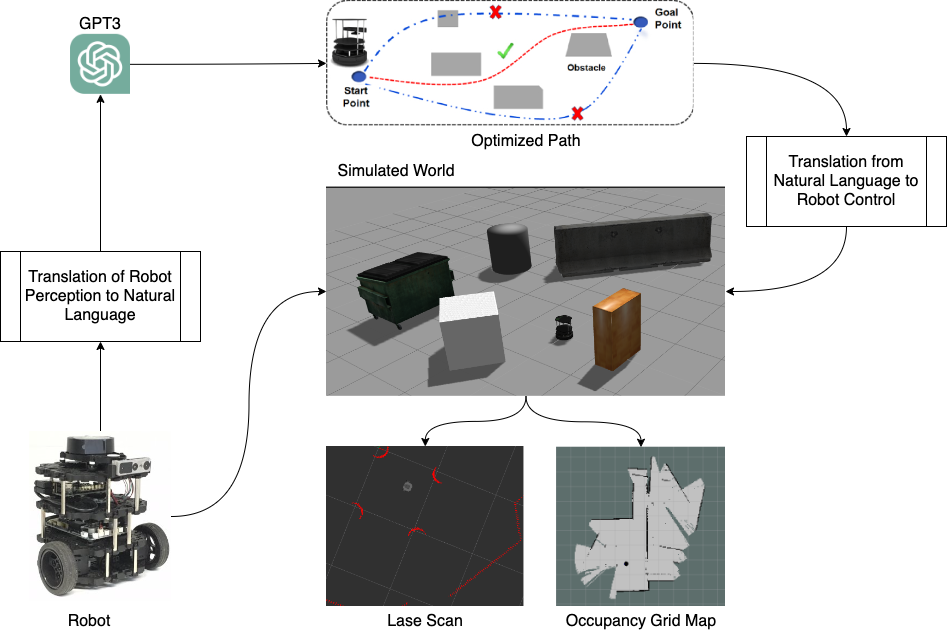}
\end{center}
\vspace{-2mm}
 \caption{Overview of the autonomous robotic path planning using LLM (GPT3.5-turbo)}
 \label{fig:overview}
 \vspace{-4mm}
\end{figure}

GPT-3.5-turbo in robotics has also shown encouraging outcomes in recent studies. The cutting-edge language model GPT-3.5, created by OpenAI, has demonstrated outstanding performance in various natural language processing tasks, including language production, question answering, and machine translation \cite{radford2019language,latif2024fine,latif2023automatic}. Due to its excellent accuracy and few-shot learning capabilities, GPT-3.5-turbo excels in these tasks, which is attributed to its capacity to identify patterns in vast amounts of data \cite{brown2020language,latif2023ai,latif2024g}. For instance, A study \cite{finnie2022robots,latif2023knowledge,lee2024applying} demonstrated that  GPT-3.5-turbo could automate the robot software's production and utilized GPT-3.5-turbo to produce code for robotic tasks in real time. In addition, GPT-3.5-turbo was shown to have the potential for human-robot interaction when it was utilized to generate natural language instructions for robotic manipulation tasks \cite{ahn2022can,lee2023gemini,lee2023multimodality}.

Our approach is motivated by assessing how well GPT-3.5-turbo applies to the robotics industry's path-planning challenge. Because it has the potential to change the sector completely, the use of GPT-3.5-turbo in robotic path planning is a fascinating topic. 
Our method converts the robotic path planning problem into a natural language problem to provide optimized paths for robotic systems. GPT-3.5-turbo may then optimize this problem to produce the desired results. The overview of the proposed approach is shown in Fig.\ref{fig:overview}.

The key contributions of the paper are:
\begin{itemize}
    \item We propose an LLM-powered robotic system for autonomous path planning.
    \item We suggest a probalistic transformation mechanism for signal to language and back to robotic motion.
    \item We evaluated the performance of proposed system with SOTA path planners (A*\cite{dang2019graph} and RRT \cite{cao2019rrt}) and observed least distance travelled and shortest planning time.
\end{itemize}

\section{Related work}
\label{sec:related-work}

Robotics is developing quickly, and with the incorporation of AI and machine learning, robots are becoming more independent and able to carry out more extensive activities. However, giving robots instructions in plain language is still challenging in the industry. Here, using LLM, such as GPT-3, can substantially contribute. This study of the literature looks at recent advances in using LLM for robotic path planning.

One of the most current publications presents a framework for correcting robot plans with natural language feedback, \cite{sharma2022correcting}. The authors demonstrate how their method can enhance robot performance while carrying out tasks based on natural language feedback from humans. Their research focuses on the problem of improving robot plans using input in natural language. However, its strategy is constrained because it primarily focuses on updating current projects rather than creating new ones.

Another fantastic piece, \cite{kuo2020deep}, shows a deep compositional robotic planner that obeys orders given in spoken language. In a series of studies, the authors prove the efficacy of their method, demonstrating that it can follow detailed instructions in plain language and produce courses ideal for robotic navigation. Robots can now comprehend imperfect natural language instructions by applying common sense reasoning thanks to another piece of work \cite{chen2020enabling}. The authors demonstrate how their method can help robots more effectively accomplish tasks based on faulty instructions. However, these activities do not consider the difficulties in developing new plans.

An intriguing method for rooting language in robotic affordances is presented in the paper \cite{ahn2022can} by Google researchers. The authors demonstrate that considering the physical affordances of the surroundings and the robot's capabilities, their technique can give robots a better grasp of natural language orders. However, their strategy is restricted to the particular pick-and-place task in simulation and does not apply to actual robotic situations.

The ability of high-level planning languages like PDDL \cite{fox2003pddl2} to root linguistic directives with symbolic representations has been extensively studied. Additionally, some methods combine task and motion planning using task-specific symbolic representations \cite{eppe2019semantics}. These methods can efficiently plan in huge continuous spaces. Still, they need a symbolic representation of the task, must be manually generated, cannot learn new concepts, and struggle with ambiguity. On top of these structures, previous research has developed models that can faithfully execute linguistic commands \cite{paul2018temporal}. However, this earlier effort cannot learn concepts that are difficult to articulate in the target planning language and cannot learn new primitives in that language due to the underlying symbolic character of the final representation. According to a study \cite{paxton2019prospection}, a model that learns to map phrases to a list of subgoals can automatically break down tasks for these planners.

A recent technique \cite{fu2019language} illustrates how multi-task reinforcement learning in a discrete state and action space may map phrases to robotic actions (navigation, pick, and place). A drone can be controlled by anticipating the robot's objective configuration \cite{blukis2018mapping}. Their model does not include object interactions, manipulations, or impediments but functions in a continuous space. It is impossible to predict a single final objective for such intricate multistep activities since the goal must also include the positions of the other items and the robot.

Our method incorporates the most recent developments in big language models, especially OpenAI's GPT-3, to build efficient robot path plans based on natural language input. To create optimized plans for the robot, we translate the robotic path planning problem into human language and apply GPT-3. We want to enhance the accuracy and effectiveness of path planning in real-world contexts by adding GPT-3.


\section{Proposed method}
\label{sec:method}
This work is inspired by the robotic perception translation framework for the pick and place task \cite{ahn2022can}. We will use a similar technique for translation and formulate a noval interface for LLM-robot interaction for efficient path planning.

Our system receives a goal as user-instruction in natural language $i$ describing the goal location in the environment. We also assume that we are given a set of actions $\mathrm{A}$, where each action $a\in \mathrm{A}$ is constrained to the robotic motion, surrounding information by Occupancy Grid Map (OGM) $m \in \mathrm{M}$ such as moving forward and taking a right turn using the mapping information, and comes with a concise language description $l_a$ (e.g., "move straight then take a right turn to reach the goal") and a utility function $P(u_a \mid m, s_o,s_g, l_A)$, which indicates the probability of $a$ for motion with description $l_a$ successfully from state $s_o$ to $s_g$ in map $m$. Intuitively, $P(u_a|m,s, l_a)$ means "if I ask the robot to do $l_a$ given map $m$ and current state $s$, will it do it?".
 
As mentioned above, $l_a$ denotes the textual label of action $a$, and $P(u_a \mid m,s, l_m)$ denotes the probability that action $a$ with textual label $l_m$ completes if executed from state $s$ in map $m$, where $u_a$ is a Bernoulli random variable. The LLM provides $P(l_a|i)$, the probability that an action's textual label is a valid next step for the user's instruction. However, we are interested in the likelihood that a given action completes the education, which we denote as $P(u_i \mid i,m, s_o,s_g,l_a)$. Assuming that an action that succeeds makes progress on $i$ with probability $P(l_a\mid i)$ (i.e., its probability of being the correct action), and an action that fails makes progress with probability zero, we can factorize this as:
\begin{equation}
    P(u_i |m,i, s_o,s_g, l_a) \propto P(u_a |m, s_o,s_g,l_a) P(l_a|i),
\end{equation}.
This corresponds to multiplying the probability of the language description of the skill given the instruction $P(l_a|m, i)$, which we refer to as task-grounding. The probability of the skill is possible in the current state of the world $P(\mathrm{u}_a|i)|s_o,s_g, l_a)$, which we refer to as world-grounding. For each state, $s$, and obtained $m$, we translate action $a$ as $l_a$ and feed this information to GPT-3.5-turbo for path planning. Later, planned path in natural langauge will be translated back to robotic action based on $P(u_i \mid i,m, s_o,s_g,l_a)$. The robot performs an action to reach $s_g$, and in case of encountering a dynamic obstacle, the robot performs a translated observation into language and receives an updated action set from GPT-3.5-turbo to achieve a goal while avoiding the obstacle. Alg.~\ref{alg:main_algorithm} presents the LLM (GPT-3.5) based Route Planning algorithm. 

\begin{algorithm}[t]
\caption{LLM (GPT-3) based Path Planning}
\label{alg:main_algorithm}
\begin{algorithmic}[1]
\State Current state $s_o$, goal state $s_g$, set of actions $A$, and their language descriptions $l_A$
\State Initialize action itself  $a=none$ and state associated $s_a$, path $P$
\While{$s_a \neq s_g$}
    \State Initialize intermediate states $S=\{s_a+up,s_a+right,s_a+left,s_a+down\}$
    \For{$s_t \in  S$ and $a \in A$}
        \State $p_a^{GPT} = p(s_t | a,l_a,m,s)$ 
        \State $p_a^{util} = p(u_a | s, l_a)$
        \State $p_a = p_a^{GPT} \times p_a^{util}$
    \EndFor
    \State $a = \argmax_{a \in A} p_a$
    \State $s_a = \argmax_{s_t \in S}(p_{a}) $, $a$ link to $s_t$
    \State $P = P \cup s_a$
    \State execute $P$ in environment, and update state $s_a$
\EndWhile
\end{algorithmic}
\end{algorithm}

\begin{figure*}
 \includegraphics[width=0.33\linewidth]{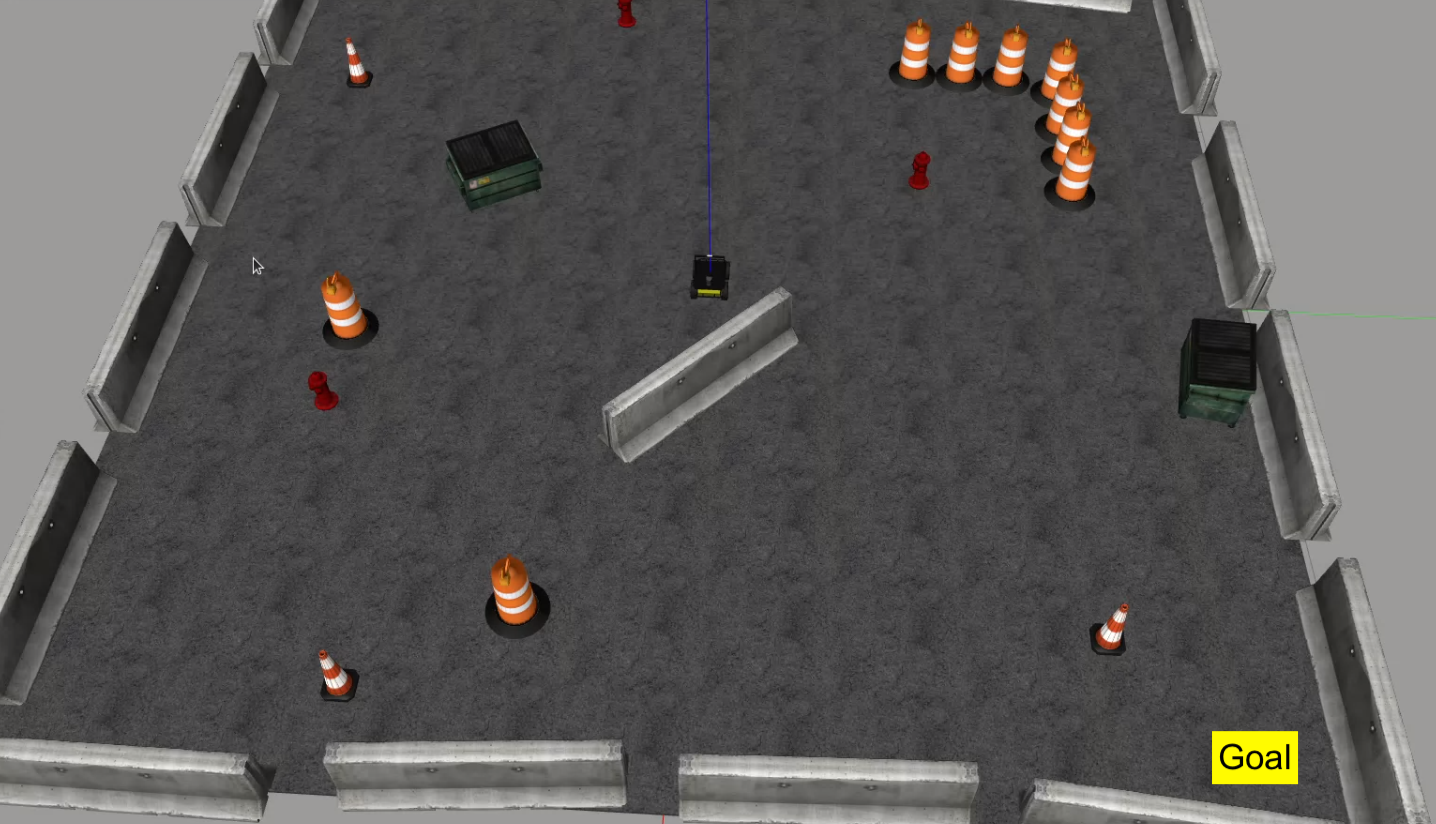}
 \label{fig:gazebo}
 \includegraphics[width=0.22\linewidth]{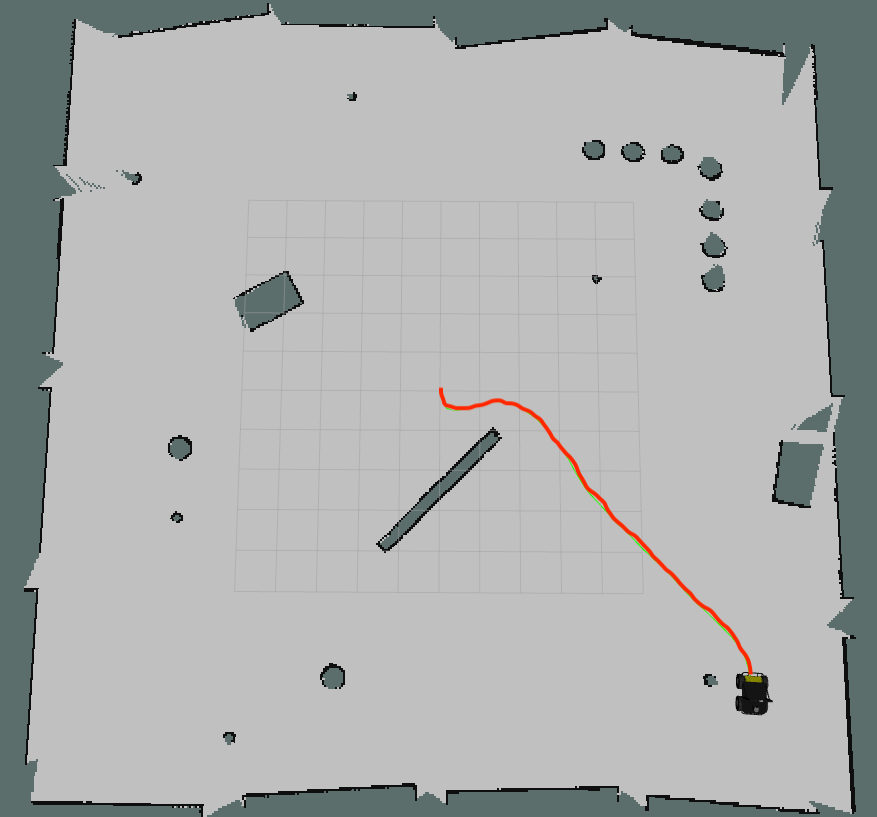}
 \label{fig:a-star}
 \includegraphics[width=0.20\linewidth]{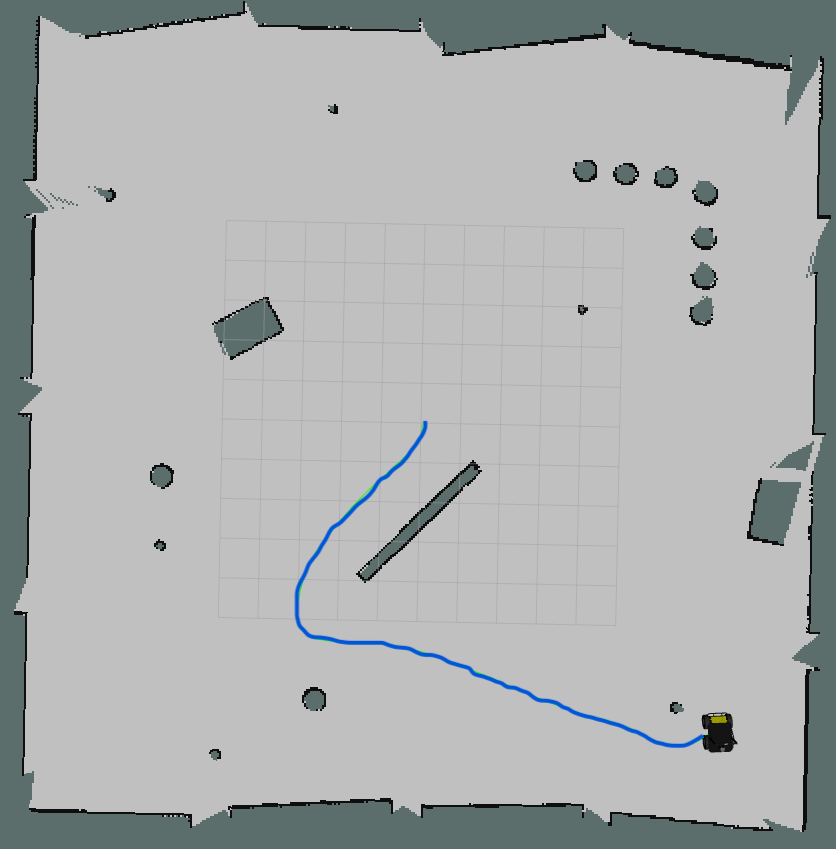}
  \label{fig:rrt}
\includegraphics[width=0.21\linewidth]{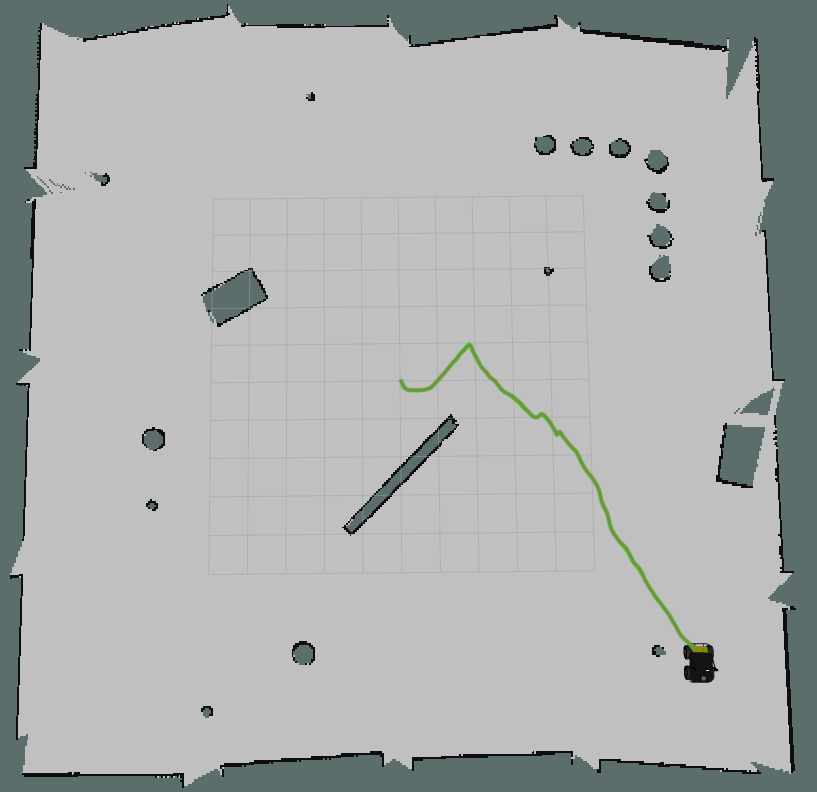}
  \label{fig:gpt}
  \vspace{-2mm}
\caption{Trajectory comparison; \textbf{Extreme Left} Gazebo World for navigation \textbf{(Left)} A*, \textbf{(Center)} RRT, and \textbf{(Right) GPt-3.5-turbo}}
 \label{fig:trajectories}
 \vspace{-4mm}
\end{figure*}

\section{Experimentation}

By putting the proposed approach to the test on several path planning tasks, including A* and RRT, comparing its results to those of other path planning algorithms already in use, and examining its computing complexity and effectiveness, we will assess the method's effectiveness. The generated paths can be compared to the ideal paths to gauge how accurate the algorithm is, and the execution time and memory consumption can be used to gauge its effectiveness.

To initiate the experimentation by providing environmental information to the LLM model using OpenAI python API\footnote{\url{https://github.com/openai/openai-python}} and utilized conversational model GPT-3.5-turbo. We set up a single robot junk world in the Gazebo simulator\footnote{\url{https://staging.gazebosim.org/home}}.
Robots are controlled through velocity commands directed by LLM responses and navigate through the environment. We observed that by default, GPT-3.5-turbo uses A* \cite{dang2019graph} algorithm for path finding from a grid, and upon prompt tuning, it changed the navigation algorithm to Hierarchical Annotated A* \cite{tellex2011understanding} for efficiency. Gazebo works on Robot Operating System (ROS)\footnote{\url{https://www.ros.org/}} and messages being published asynchronously. The robot is equipped with a Lidar and uses odometry data to determine current position. We use GMapping \cite{balasuriya2016outdoor} to create a map using scan data and provide it to the language model as environment information. We created an intermediate service as a translator to integrate robotic motion into the GPT, which subscribes to the robot position, scans data from Lidar, maps data, and converts it into a natural language that uses a language model (GPT-3.5-turbo) as a prompt. The language model provides coordination in a structured language, parsed into a list of coordinates for navigation. Intermediate service created velocity commands to navigate the robot through the generated coordinates and publish to the ROS Core for robot motion. We have performed multiple trials and observed consistent trajectories generated by GPT-3.5-turbo. To evaluate the performance, we also ran RRT and A* and reported \textbf{Processing time}, \textbf{path correctness}, and \textbf{Path length}.

\begin{figure*}
 \includegraphics[width=0.33\linewidth]{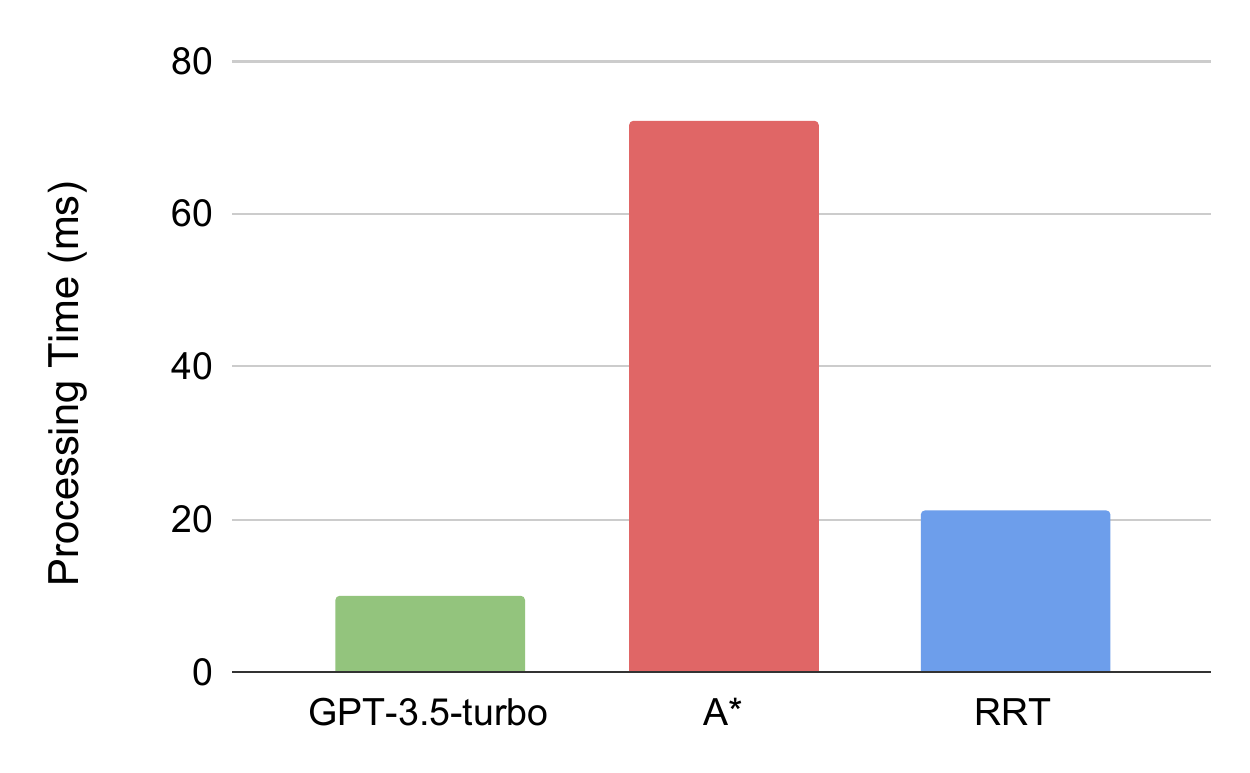}
 \label{fig:time}
 \includegraphics[width=0.33\linewidth]{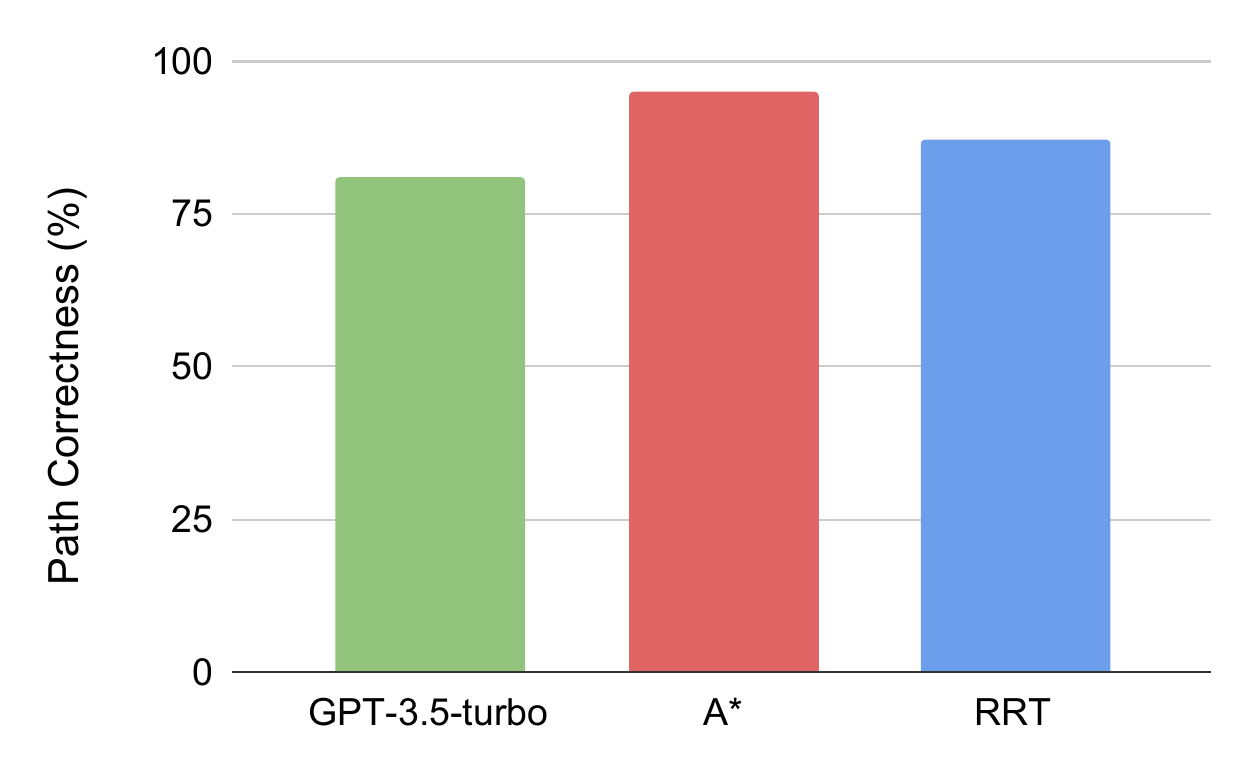}
  \label{fig:correctness}
\includegraphics[width=0.33\linewidth]{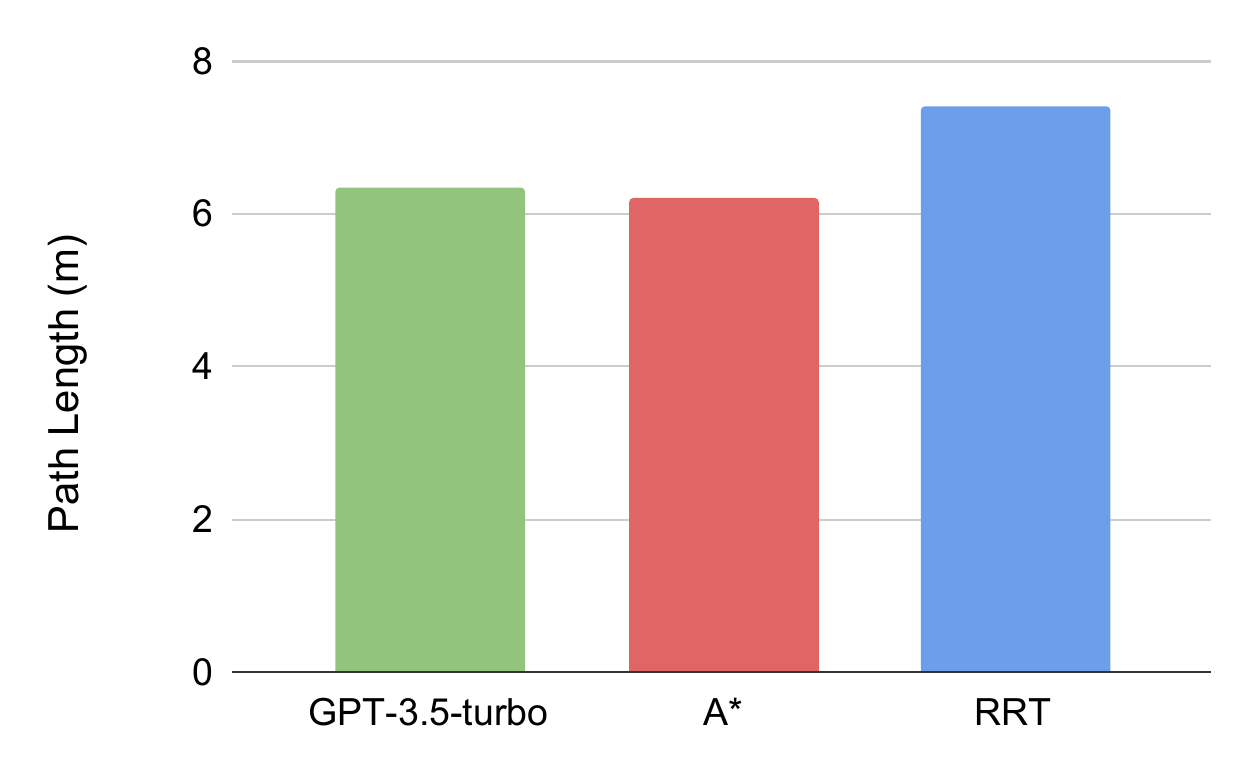}
  \label{fig:length}
  \vspace{-2mm}
\caption{Performance compasrion plots; \textbf{(Left)} Processing Time, \textbf{(Center)} Path Correctness, and \textbf{(Right) Path Length}}
 \label{fig:results}
 \vspace{-4mm}
\end{figure*}

\section{Results}
In the experiment, the processing time, path correctness, and path length of GPT-3.5-turbo were compared to those of two other algorithms, A* and RRT. GPT-3.5-turbo displayed a remarkable processing time of 10ms, as shown in Figure~\ref{fig:results}, much less than that of A* (72ms) and RRT. (21ms). In addition, GPT-3.5-turbo obtained an average path length of 6.34m and a path accuracy rate of 81\%.

Although GPT-3.5-turbo's path correctness was lower than that of A* (95\%) and RRT (87\%), its superiority can be attributed to the processing time, which was much faster. This shorter processing time is most likely a result of GPT-3.5-turbo's quick analysis and generation of optimal pathways thanks to its effective language comprehension capabilities. However, there is still space for improvement, as evidenced by the lower accuracy rate and slightly longer path length. The A* algorithm, on the other hand, is renowned for its great accuracy, but its longer processing time is a drawback. RRT processes information more quickly than A*, but at the cost of a lower path accuracy and a longer average path length, two of its primary drawbacks. Figure~\ref{fig:trajectories} has shown the gazebo world of simulation along with the trajectories followed bye ach algorithm.

As a result of its short processing time, GPT-3.5-turbo shows promise for real-time applications, but additional advancements in path correctness and length are required for it to be regarded as the best option. Even while A* and RRT both have advantages of their own, in challenging situations, GPT-3.5-turbo is more effective and versatile. When selecting the best pathfinding algorithm, it is crucial to consider the specific requirements of an application.

\section{Discussion}
Due to its quick processing time, which is necessary for real-time applications, GPT-3.5-turbo has demonstrated potential as a revolutionary technique for robot navigation. GPT-3.5-turbo can swiftly assess complicated settings and produce effective pathways using its potent language understanding skills. GPT-3.5-turbo has its limits in terms of accuracy and optimality, as evidenced by the lower path correctness rate and marginally longer path length compared to conventional algorithms like A* and RRT. Furthermore, GPT-3.5-turbo extensively uses natural language understanding, which may not necessarily be the best strategy for essentially geometric issues, such as robot navigation.

Combining classical algorithms with GPT-3.5-turbo may be an option to enhance the performance of GPT-3.5-turbo for robot navigation. By combining the advantages of both approaches, this hybrid strategy might benefit from the quick processing time of the GPT-3.5-turbo and the improved accuracy of algorithms like A*. Additionally, including extra data could give GPT-3.5-turbo the context it needs to produce more precise and ideal pathways. Additional data might include extensive environment maps and real-time sensor information. Last, the model can increase its knowledge and pathfinding abilities with repeated training on various navigation circumstances.

\section{Conclusion}
In real-time applications where processing speed is crucial, the GPT-3.5-turbo offers a potential alternative for robot navigation. However, its path correctness and length shortcomings draw attention to the need for advancements and other strategies to be considered. The optimum answer might be a hybrid strategy combining conventional algorithms' benefits with the quick processing speed of GPT-3.5-turbo. More study and development are necessary to maximize the GPT-3.5-turbo's performance and fully tap into its potential for robot navigation applications.




\bibliographystyle{IEEEtran}
\bibliography{refs}

\end{document}